# Conditional Probability Tree Estimation Analysis and Algorithms

**Alina Beygelzimer**
IBM Research
beygel@us.ibm.com

**John Langford**
Yahoo! Research
jl@yahoo-inc.com

**Yuri Lifshits**
Yahoo! Research
yury@yury.name

**Gregory Sorkin**
IBM Research
sorkin@us.ibm.com

**Alex Strehl**
Yahoo! Research
astrehl@gmail.com

## Abstract

We consider the problem of estimating the conditional probability of a label in time $O(\log n)$, where $n$ is the number of possible labels. We analyze a natural reduction of this problem to a set of binary regression problems organized in a tree structure, proving a regret bound that scales with the depth of the tree. Motivated by this analysis, we propose the first online algorithm which provably constructs a logarithmic depth tree on the set of labels to solve this problem. We test the algorithm empirically, showing that it works succesfully on a dataset with roughly $10^6$ labels.

## 1 Introduction

The central question in this paper is how to efficiently estimate the conditional probability of label $y \in \{1, \ldots, n\}$ given an observation $x \in X$. Virtually all approaches for solving this problem require $\Omega(n)$ time. A commonly used one-against-all approach, which tries to predict the probability of label $i$ versus all other labels, for each $i \in \{1, \ldots, n\}$, requires $\Omega(n)$ time per training example. Another common $\Omega(n)$ approach is to learn a scoring function $f(y, x)$ and convert it into a conditional probability estimate according to $f(y, x)/Z(x)$, where $Z(x) = \sum_i f(i, x)$ is a normalization factor.

The motivation for dealing with the computational difficulty is the usual one—we want the capability to solve otherwise unsolvable problems. For example, one of our experiments involves a probabilistic prediction problem with roughly $10^6$ labels and $10^7$ examples, where any $\Omega(n)$ solution is intractable.

### 1.1 Main Results

In Section 4, we provide the first online supervised learning algorithm that trains and predicts with $O(\log n)$ computation per example. The algorithm does not require knowledge of $n$ in advance; it adapts naturally as new labels are encountered.

The prediction algorithm uses a binary tree where regressors are used at each node to predict the conditional probability that the true label is to the left or right. The probability of a leaf is estimated as the product of the appropriate conditional probability estimates on the path from root to leaf. In our experiments, we use linear regressors trained via stochastic gradient descent.

The difficult part of this algorithm is constructing the tree itself. When the number of labels is large, it becomes critical to construct easily solvable binary problems at the nodes. In Section 4.2, we introduce a tree-construction rule with two desirable properties. First, it always results in depth $O(\log n)$. It also encourages natural problems by minimizing expected loss at the nodes. The technique used in the algorithm is also useful for other prediction problems such as multiclass classification.

We test the algorithm empirically on two datasets (in Section 4.3), and find that it both improves performance over naive tree-building approaches and competes in prediction performance with the common one-against-all approach, which is exponentially slower.

Finally, we analyze a broader set of logarithmic time probability estimation methods. In Section 3.1 we prove that any tree based approach has squared loss bounded by the tree depth squared times the average squared loss of the node regressors used. In contrast, the PECOC approach [4] has squared loss bounded by just 4 times the average squared loss but uses $\Omega(n)$ computation. This suggests a tradeoff between computation and squared loss multiplier. Section 3.2 de-scribes a $k$-parameterized construction achieving a ratio of $4(\log_k n)^2 \left(\frac{k-1}{k}\right)^2$ while using $O(k \log_k n)$ computation, where $k = 2$ gives the tree approach and $k = n$ gives PECOC.

## 1.2 Prior Work

There are many methods used to solve conditional probability estimation problems, but very few of them achieve a logarithmic dependence on $n$. The ones we know are batch constructed regression trees, C4.5 [9], ID3 [7], or Treenet [10], which are both too slow to consider on datasets with the scale of interest, and incapable of reasonably dealing with new labels appearing over time.

Mnih and Hinton [8] constructed a special purpose tree-based algorithm for language modeling, which is perhaps the most similar previous work. The algorithm there is specialized to word prediction and is substantially slower since it involves many iterations through the training data. However, the general analysis we provide in Section 3.1 applies to their algorithm. We regard the empirical success of their algorithm as further evidence that tree-based approaches merit investigation.

## 1.3 Outline

Section 3 states and analyses methods for logarithmic time probabilistic prediction given a tree structure. Section 4 gives an algorithm for building the tree structure. The analysis in the first section is sufficiently general so that it applies to the second.

## 2 Problem Setting

Given samples from a distribution $P$ over $X \times Y$, where $X$ is an arbitrary observation space and $Y = \{1, \ldots, n\}$, the goal is to estimate the conditional probability $P(y \mid x)$ of a label $y \in Y$ for a new observation $x \in X$.

For an estimator $Q(y \mid x)$ of $P(y \mid x)$, the *squared loss* of $Q$ with respect to $P$ is defined as

$$\ell_P(Q) = \mathbf{E}_{(x,y) \sim P}(P(y \mid x) - Q(y \mid x))^2. \quad (1)$$

It is more common to define an observable squared loss where $P(y \mid x)$ in equation (1) is replaced by 1. We consider *regret* with respect to the common definition, since it is well known that the difference between observable squared loss and the minimum possible observable squared loss is equal to $\ell_P(Q)$. We therefore use regret and squared loss interchangeably in this paper.

It is well known that squared loss is a strictly proper scoring rule [2], thus $\ell_P(Q)$ is uniquely minimized by $Q = P$. Our analysis focuses on squared loss because it is a bounded proper scoring rule. The boundedness implies that convergence guarantees hold under weaker assumptions than for unbounded proper scoring rules such as log loss.

## 3 Probabilistic Prediction Given a Tree

This section assumes that a tree structure is given, and analyzes how to use it for probabilistic logarithmic time prediction.

### 3.1 Conditional Probability Tree

Consider a fixed binary tree whose leaves are the $n$ labels. For a leaf node $y \in Y$, let $T(y)$ be the set of non-leaf nodes on the path from the root to $y$ in the tree.

Each non-leaf node $i$ is associated with the regression problem of predicting the probability, under $P$, that the label $y$ of a given observation $x \in X$ is in the left subtree of $i$, conditioned on $i \in T(y)$. The following procedure shows how to transform multiclass examples into binary examples for each non-leaf node in the tree. Here $\text{right}_i(y)$ is 0 when $y$ is in the left subtree of node $i$, and 1 otherwise.

---

**Algorithm 1**: Conditional Probability Tree Training (training set $S$, regression algorithm $R$)

---

**foreach** internal node $i$ **do**
  $S_i \leftarrow \emptyset$
**foreach** example $(x, y) \in S$ **do**
  **foreach** node $i \in T(y)$ **do**
    Add $(x, \text{right}_i(y))$ to $S_i$.
**foreach** internal node $i$ **do**
  train $f_i = R(S_i)$

---

Given a new observation $x \in X$ and a label $y \in Y$, we use the learned binary regressors $f_i$ to estimate $P(y \mid x)$. Letting $Q_i(1 \mid x) = f_i(x)$ and $Q_i(0 \mid x) = 1 - f_i(x)$, we define the estimate

$$Q(y \mid x) = \prod_{i \in T(y)} Q_i(\text{right}_i(y) \mid x). \quad (2)$$

#### 3.1.1 Analysis of the Conditional Probability Tree

Algorithm 1 implicitly defines a distribution $P_i$ over $X \times \{0, 1\}$ induced at node $i$: A sample from $P_i$ is obtained by drawing $(x, y)$ according to $P$ until $i \in T(y)$, and outputting $(x, \text{right}_i(y))$ (although we never explicitly perform this sampling). The following theorem bounds the squared loss of $Q$ given the average squared loss of the binary regressors.

**Theorem 1.** *For any distribution $P$, any set of node estimators $Q_i$, and any pair $(x, y)$, with $Q$ given by equation (2),*

$$(Q(y \mid x) - P(y \mid x))^2$$
$$\leq d^2 \, \mathbf{E}_i \left( Q_i(\text{right}_i(y) \mid x) - P_i(\text{right}_i(y) \mid x) \right)^2,$$

*where $d = |T(y)|$ and the expectation is over $i$ chosen uniformly at random from $T(y)$.*

*Proof.* We use Lemma 2. Using the notation of its proof, observe that

$$\left( \sum_{i=1}^d |q_i - p_i| \right)^2 = d^2 \left( \mathbf{E}_i \, |q_i - p_i| \right)^2$$
$$\leq d^2 \mathbf{E}_i \left( |q_i - p_i|^2 \right)$$

using Jensen's inequality. □

Most of the theorem is proved with the following core lemma. For a node $i$ on the path from the root to label $y$, define $p_i = P_i(\text{right}_i(y) \mid x)$, the conditional probability that the label is consistent with the next step from $i$ given that all previous steps are consistent. Similarly define $q_i = Q_i(\text{right}_i(y) \mid x)$.

**Lemma 2.** *For any distribution $P$, any set of node estimators $Q_i$, and any pair $(x, y)$, with $Q$ given by equation (2),*

$$|Q(y \mid x) - P(y \mid x)| \leq \sum_{i \in T(y)} |q_i - p_i| \prod_{j \neq i} \max\{p_j, q_j\}$$
$$\leq \sum_{i \in T(y)} |q_i - p_i|.$$

The last inequality is the simplest—it says the differences in errors add. However, the quantity after the first inequality can be much tighter.

*Proof.* We first note that

$$|Q(y \mid x) - P(y \mid x)| \leq \prod_i \max\{p_i, q_i\} - \prod_i \min\{p_i, q_i\}$$

since $\prod_i \max\{p_i, q_i\} \geq \max\{Q(y \mid x), P(y \mid x)\}$ and $\prod_i \min\{p_i, q_i\} \leq \min\{Q(y \mid x), P(y \mid x)\}$.

We use a geometric argument. With $\prod_i \min\{p_i, q_i\}$ defining the volume of one "corner" of a cube with sides $\max\{p_i, q_i\}$, slabs $|q_i - p_i| \prod_{j \neq i} \max\{p_j, q_j\}$ fill in the remaining volume (with overlap). Consequently, we can bound the difference in volume as

$$\prod_i \max\{p_i, q_i\} - \prod_i \min\{p_i, q_i\}$$
$$\leq \sum_i |q_i - p_i| \prod_{j \neq i} \max\{p_j, q_j\}$$
$$\leq \sum_i |q_i - p_i|,$$

since all $p_j$ and $q_j$ are bounded by 1. □

As suggested by the proof, the lemma's bound can be asymptotically tight. If all $p_i$ are equal to some $p$ and all $|q_i - p_i|$ are small, the left side is approximately $p^{d-1} \sum_i |q_i - p_i| = dp^d \mathbf{E} |q_i - p_i|$, a factor $p^d$ times the right side.

### 3.2 Conditional PECOC

The conditional probability tree is as computationally tractable as we could hope for, but is not as robust as we could hope for. For example, the PECOC approach [4] yields a squared loss multiplier of 4 independent of the number of labels. Is there an approach more robust than the tree, but requiring less computation than PECOC?

We provide a construction which trades off between the extremes of PECOC and the conditional probability tree. The essential idea is to shift from a binary tree to a $k$-way tree, where PECOC with $k - 1$ regressors is used at each node in the tree to estimate the probability of any child conditioned on reaching the node. For simplicity, we assume that $k$ is a power of 2, and $n$ is a power of $k$.

**Theorem 3.** *Pick a $k$-way tree on the set of $n$ labels, where $k$ is a power of 2. For all distributions $P$ and all sets of learned regressors, with $k - 1$ regressors per node of the tree, for all pairs $(x, y)$,*

$$(Q(y \mid x) - P(y \mid x))^2 \leq 4(\log_k n)^2 \left( \frac{k-1}{k} \right)^2 \epsilon^2,$$

*where $\epsilon^2$ is the average squared loss of the $(k-1) \log_k n$ questioned regressors.*

*Proof.* The proof is by composition of two lemmas.

In each node of the tree, Lemma 4 bounds the power of the adversary to disturb the probability estimate as a function of the adversary's regret. Similarly, Lemma 2 bounds the power of the adversary to induce an overall misestimate as a function of the adversary's power to disturb the estimates within each node on the path. □

The curve below illustrates how the construction trades off computation for a better regret bound as a function of $k$.

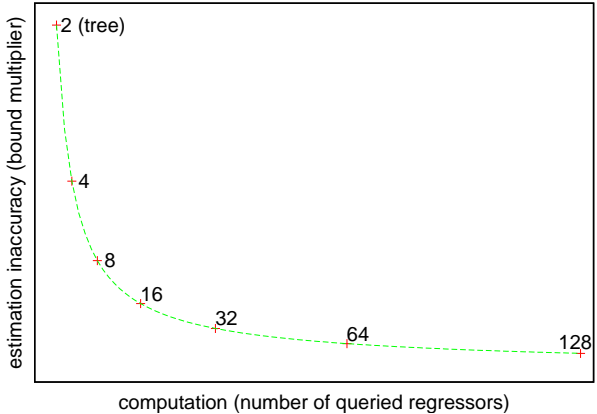

To complete the proof of Theorem 3 we describe the PECOC construction in Section 3.2.1 and prove Lemma 4 in Section 3.2.2.

### 3.2.1 The PECOC Construction

The PECOC construction is defined by a binary matrix $C$ with each column a label and each row defining a regression problem. The regression problem corresponding to row $i$ is to predict the probability given $x$ that the correct label is in the subset

$$Y_i = \{y \in Y : C(i,y) = 1\}. \qquad (3)$$

We use an explicit family of Hadamard codes given by the recursive formula

$$C_2 = \begin{bmatrix} 1 & 1 \\ 1 & 0 \end{bmatrix}, \quad C_{2^t} = \begin{bmatrix} C_t & C_t \\ C_t & 1 - C_t \end{bmatrix}.$$

We use a matrix $C_{2^t}$ with $2^{t-1} \leq n < 2^t$, noting that its size $2^t$ is less than $2n$; if $2^t > n$ we simply add dummy labels. We henceforth assume without loss of generality that $n$ is a power of 2. We train PECOC according to the following algorithm.

---
**Algorithm 2**: PECOC Training (training set $S$, regression algorithm $R$)

**for** each row $i$ of $C$ **do**
  Let $S_i = \{(x, C(i,y)) : (x,y) \in S\}$
  **train** $r_i = R(S_i)$.
---

Given a new observation $x \in X$ and a label $y \in Y$, PECOC uses the binary regressors $r_i$ learned in Algorithm 2 to estimate $P(y \mid x)$ using the formula

$$\text{pecoc}(y \mid x) = 2\,\mathbf{E}_i\big[C(i,y)r_i(x) + (1 - C(i,y))(1 - r_i(x))\big] - 1, \quad (4)$$

where the expectation is over $i$ drawn uniformly from the rows of $C$. The reason for this formula is clarified by the proof of Lemma 4.

### 3.2.2 A Careful PECOC analysis

The following theorem gives the precise regret bound, which follows from the analysis in [4] but is tighter for small values of $n$ than the bound stated there.

**Lemma 4.** (PECOC regret [4]) For all distributions $P$ and all sets of regressors $r_i$ (as defined in Algorithm 2), for all $x \in X$ and $y \in Y$,

$$(\text{pecoc}(y \mid x) - P(y \mid x))^2 \leq$$
$$4\left(\frac{n-1}{n}\right)^2 \mathbf{E}_i(r_i - P(y \in Y_i \mid x))^2,$$

where $Y_i$ is the subset defined by row $i$ per (3).

*Proof.* Since the code and the prediction algorithm are symmetric with respect to set inclusion, we can assume without loss of generality that $y$ is in every subset (complementing all subsets not containing $y$). Thus every entry $C(i,y) = 1$, and by (4) the PECOC output estimate of $P(y \mid x)$ is

$$\text{pecoc}(y \mid x) = \frac{2}{n}\sum_{i=1}^{n} r_i(x) - 1.$$

Let $\bar{r}_i(x) = P(y \in Y_i \mid x) = \sum_{v \in Y_i} P(v \mid x)$ denote the perfect subset estimators, and write $r_i(x) = \bar{r}_i(x) + \epsilon_i$. By the nature of $C$, the label $y$ under consideration occurs in every subset, and every other label $v \neq y$ in exactly half the subsets, so that

$$\sum_i r_i(x) = \sum_i \left(\sum_{v \in Y_i} P(v \mid x) + \epsilon_i\right)$$
$$= \sum_v \sum_{i:\, Y_i \ni v} P(v \mid x) + \sum_i \epsilon_i$$
$$= \sum_{v \neq y} \frac{n}{2} P(v \mid x) + nP(y \mid x) + \sum \epsilon_i$$
$$= \frac{n}{2}(1 + P(y \mid x)) + \sum \epsilon_i.$$

This gives $\text{pecoc}(y \mid x) = P(y \mid x) + \frac{2}{n}\sum_i \epsilon_i$, for squared loss $(\text{pecoc}(y \mid x) - P(y \mid x))^2 = (\frac{2}{n}\sum_i \epsilon_i)^2$. One of the subsets, say the first, is trivial (it includes all labels), and for it we stipulate the true probability $r_1 = 1$, so $\epsilon_1 = 0$. Letting $\mathbf{E}_i \epsilon_i$ denote the mean of the other $n-1$ errors $\epsilon_i$, the squared loss is $(2\frac{n-1}{n}\mathbf{E}_i\epsilon_i)^2$, establishing the theorem. □

# 4 Online Tree Construction

The analysis of Section 3.1.1 applies to any binary tree, and motivates the creation of trees which have small depth and small regret at the nodes. This leaves the question, "Which tree should we use?" We give an online tree construction algorithm with several useful properties. In particular, the algorithm doesn't require any prior knowledge of the labels, and takes $O(\log n)$ computation per example, when there are $n$ labels. The algorithm guarantees a tree with $O(\log n)$ maximum depth using a decision rule that trades off between depth and ease of prediction.

## 4.1 Online Tree Building Algorithm

Algorithm 3 builds and maintains a tree, whose leaves are in one-to-one correspondence with the labels seen so far. Each node $i$ in the tree is associated with a regressor $f_i : X \to [0, 1]$. Given a new sample $(x, y) \in X \times Y$, we consider two cases.

If $y$ already exists as a label of some leaf in the tree, then there is an associated root-to-leaf path and we can use the conditional probability tree algorithms of the previous section to train and test on $(x, y)$, with one minor modification when training: we add a regressor at the leaf and train it with the example $(x, 0)$.

If $y$ does not exist in the tree, then the algorithm still traverses the tree to some leaf $j$, using a decision rule that computes a direction (left or right) at each non-leaf node encountered. Once leaf $j$ is reached, it necessarily corresponds to some label $y' \neq y$. We convert $j$ to a non-leaf node with left child $y'$ and right child $y$. The regressor at node $j$ is duplicated for $y'$. A new regressor is created for $y$ and trained on the example $(x, 0)$.

We now describe the decision rule used to decide which way to go (left or right) at each non-leaf node $i$ encountered during the traversal. First, let $L_i$ denote the number of children to the left of node $i$, and $R_i$ the number to the right. If $f_i(x) > 1/2$, where $f_i(x)$ is the current prediction associated with node $i$ on $x$, then the regressor favors the right subtree for this input, and otherwise the left subtree. If the regressor favors the side with the smaller number of elements, then this direction is chosen. If the regressor favors the side with more elements, then the algorithm faces a dilemma. On one hand, sending the new label to the right would result in a more highly balanced tree, but on the other hand it would result in a training sample disagreeing with the current regressor's prediction. Our resolution is to define an objective function

$$\text{obj}(p, L, R, \alpha) = (1 - \alpha)2(p - \tfrac{1}{2}) + \alpha \log_2 \tfrac{L}{R}$$

---

**Algorithm 3**: Online conditional probability tree (CPT) Training (regression algorithm $R$, aggressiveness $\alpha$)

---
**create** the root node $r$
**foreach** example $(x, y)$ **do**
　**if** $y$ has been seen previously **then**
　　For each $i \in T(y)$, train $f_i$ with $(x, \text{right}_i(y))$.
　**else**
　　Set $i = r$.
　　**while** $i$ is not a leaf **do**
　　　**if** $\text{obj}(f_i(x), L_i, R_i, \alpha) > 0$ **then** $c = 1$ (right)
　　　**else** $c = 0$ (left)
　　　Train $f_i$ with example $(x, c)$
　　　Set $i$ to the child of $i$ corresponding to $c$
　　Create children of leaf $i$:
　　　　left with a copy of $i$ (including $f_i$),
　　　　right with label $y$ trained on $(x, 0)$.
　　Train $f_i$ with $(x, 1)$.

---

and send the label to the right of node $i$ if

$$\text{obj}(f_i(x), L_i, R_i, \alpha) > 0. \qquad (5)$$

Here $\alpha$ is a free parameter set for the run of the entire algorithm. When $\alpha = 1$, the rule indicates that we should place new labels on the side with fewer current labels, resulting in a perfectly balanced tree. When $\alpha = 0$, the direction chosen is always the one currently favored by the regressor. A trade-off between these two objectives is provided by values of $\alpha$ between these two extremes.

Pseudo-code is provided in Algorithm 3.

## 4.2 Online Tree Building Analysis

In this section we analyze Algorithm 3. Throughout the section, for any tree node under consideration, we will use $N$ for the total number of leaves under the node, $L$ the number on the left and $R$ on the right, with $L + R = N$. We note that rule (5) is symmetric with respect to $L$ and $R$. We also define

$$\kappa = \frac{1}{1 + 2^{1-1/\alpha}}.$$

Claim (6) will establish that at most about a fraction $\kappa$ of the leaves can fall on either side of a node, with $\kappa = 1/2$ for $\alpha = 1$ and $\kappa \to 1$ as $\alpha \to 0$.

**Claim 5.** *If a node has $L$ leaves in its left subtree, $R$ in the right, and $N = L + R$ altogether, if $R/N > \kappa$ then a new leaf is added to the left subtree regardless of the prediction value $p$ at the node (and symmetrically for $L$).*

*Proof.* For any $p \in [0,1]$,

$$\text{obj}(p, L, R, \alpha) \leq (1-\alpha)2(1-\tfrac{1}{2}) - (1-\alpha)$$
$$= (1-\alpha) + \alpha \log_2 \tfrac{L}{R},$$

which is $< 0$ (forcing a leaf to be added to the left) if $L/R < 2^{\frac{\alpha-1}{\alpha}}$, or equivalently if $R/N > \kappa$. □

**Claim 6.** *Under any non-leaf node, $L, R < \kappa N + (1-\kappa)$.*

*Proof.* We prove this inductively for $R$; the result for $L$ follows symmetrically. A non-leaf node starts with one left and one right child, and $R = L = 1$, $N = 2$ satisfies the claim. Given that $R$, $L$, and $N$ satisfy the claim, we now prove that when a leaf is added, so do the next values $R'$ (either $R$ or $R+1$), $L'$ (respectively $L+1$ or $L$), and $N' = N+1$. There are two cases. If $R < \kappa N$ then

$$R' \leq R + 1 < \kappa N + 1 = \kappa(N'-1) + 1 = \kappa N' + 1 - \kappa.$$

If $R \geq \kappa N$ then the next addition is to $L$ not $R$, and

$$R' = R \leq \kappa N + 1 - \kappa < \kappa N' + 1 - \kappa.$$

□

**Theorem 7.** *For all regressors at the nodes of the tree, for all learning problems on $n$ labels, for all $\alpha \in (0,1]$ the depth of the tree is at most $\log n / \log \kappa + 2$.*

*Proof.* If the root node has $n$ leaves below it, then by the preceding claim a child ("depth 1") of the root has at most $\kappa n + (1-\kappa)$ leaves, a grandchild has at most $\kappa^2 n + \kappa(1-\kappa) + (1-\kappa)$ leaves, and a depth-$d$ child has at most

$$\kappa^d n + \kappa^{d-1}(1-\kappa) + \cdots + k(1-\kappa) + (1-\kappa) \leq \kappa^d n + 1$$

leaves, using $\sum_{d=0}^{\infty} \kappa^d = 1/(1-\kappa)$. With $d = -\lceil \ln n / \ln \kappa \rceil$, a depth-$d$ child has at most 2 leaves, and thus further depth one, and we add one more to account for the ceiling function. □

**Definition 8.** *A* disagreement *is the event when a new label reaches a node, and the algorithm decides to insert it in the subtree that is not preferred by the regressor.*

That is, a disagreement occurs when the regressor's prediction is at most $1/2$ and the label is inserted to the right, or when the prediction is greater than $1/2$ and the label is inserted to the left.

Note that the number of disagreements incurred when adding a new label (leaf) is at most the depth of that leaf, and as the tree evolves the "same" leaf (per the copying rule of the algorithm) may become deeper but never shallower. Thus the total number of disagreements incurred in building a tree is at most the sum of the depths of all leaves of the final tree.

To get a grasp on this quantity, for simplicity we disregard the additive $1 - \kappa$ in Claim 6 coming from adding vertices discretely, one at a time. (The effect is most dramatic when a node has just two children, $L = R = 1$, and adding a leaf necessarily produces a lopsided tree with $L = 1$ and $R = 2$ or vice-versa. For large values of $L + R = N$ the effect of discretization is negligible.)

As usual, for a node in a tree let $L$ be the number of leaves in its left subtree, $R$ in the right, $N = L + R$.

**Theorem 9.** *Let $T$ be an $n$-leaf binary tree in which for each node, $L, R \leq \kappa N$. Then the total of the depths of the leaves of $T$ is at most $d(n) = n \log n / H(\kappa)$, where $H(\kappa) = -\kappa \log \kappa - (1-\kappa) \log(1-\kappa)$.*

*Proof.* The proof is by induction on $n$, starting from the base case $n = 2$ where the total of the depths (or total depth for short) is 2. It is well known that the entropy function $H(\kappa)$ is maximized by $H(1/2) = \log 2$, so in the base case we do indeed have $2 \leq d(n)$ since $d(n) \geq 2 \log 2 / \log 2 = 2$.

Proceeding inductively, the total depth for an $N$-leaf tree with $L$- and $R$-leaf subtrees is the total depth of $L$ (at most $d(L)$), plus the total depth of $R$ (at most $d(R)$), plus $N$ (since each leaf is 1 deeper in the full tree). Since $d(\cdot)$ is a convex function, the worst case comes from the most unequal split, and applying the inductive hypothesis, the total depth for $N$ is at most

$$N + d(\kappa N) + d((1-\kappa)N)$$
$$\leq N + \frac{\kappa N \log(\kappa N)}{H(\kappa)} + \frac{(1-\kappa)N \log((1-\kappa)N)}{H(\kappa)}$$
$$= N + \frac{N}{H(\kappa)}(\kappa \log \kappa + \kappa \log N$$
$$\qquad + (1-\kappa)\log(1-\kappa) + (1-\kappa)\log N)$$
$$= N + \frac{N}{H(\kappa)}(-H(\kappa) + \log N)$$
$$= N \log N / H(\kappa)$$
$$= d(N),$$

completing the proof that $d(N)$ is an upper bound. □

### 4.3 Experiments

We conducted experiments on two datasets. The purpose of the first experiment is to show that the conditional probability tree (CPT) competes in prediction performance with existing exponentially slower approaches. To do this, we derive a label probability prediction problem from the publicly available Reuters

RCV1 dataset [6]. The second experiment is a full-scale test of the system where an exponentially slower approach is too intractable to seriously consider. We use a proprietary dataset that consists of webpages and associated advertisements, where the derived problem is to predict the probability that an ad would be displayed on the webpage.

Each dataset was split into a training and test set. Each training or test sample is of the form $(x, y)$. The algorithms train on the training set and produce a probabilistic rule $f(\cdot, \cdot)$ that maps pairs of the form $(x, y)$ to numbers in the range $[0, 1]$, where we interpret $f(x, y)$ as an approximation to $P(y \mid x)$. The algorithms are evaluated on the test set by computing the empirical squared loss, $\sum_{(x,y)} (1 - f(x, y))^2$. The algorithms are allowed to continue learning as they are tested, however the predictions $f(x, y)$ used above are computed before training on the sample $(x, y)$. This type of evaluation is called "progressive validation" [1] and accurately measures the performance of an online algorithm. In particular, it is an unbiased estimate of the algorithm's performance under the assumption that the $(x, y)$ pairs are identically and independently distributed. In the motivating applications of our algorithm, we expect new labels to appear throughout the learning process, which requires learning to occur continually in an online fashion. Thus, turning learning off and computing a "test loss" is less natural. Nevertheless, for the Reuters dataset, we verified that the test loss and progressive validation are quite similar. For the web advertising dataset, the two measures were drastically different (all methods performed much worse under test loss), due to the large number of labels that appear only in the test set.

The CPT algorithm was executed with three tree-building construction methods: a random tree where uniform random left/right decisions were made until a leaf was encountered, a balanced tree according to algorithm 3 with $\alpha = 1$, and a general tree according to algorithm 3 with $\alpha < 1$. For the binary regression problems (at the nodes), we used Vowpal Wabbit [5], which is a simple linear regressor trained by stochastic gradient descent. One essential enabling feature of VW is a hashing trick (described in [11, 12]) which allows us to represent $1.7M$ linear regressors on a sparse feature space in a reasonable amount of RAM.

### 4.3.1 Reuters RCV1

The Reuters dataset consists of about $800K$ documents, each assigned to one or more categories. A total of approximately 100 categories appear in the data. We split the data into a training set of $780K$ documents and a test set of $20K$ documents, opposite to its original use. For each document doc, we formed an example of the form $(x, y)$, as follows. The vector $x$ uses a "bag of words" representation of doc, weighted by the normalized TF-IDF scores, exactly as done in the paper [6]. The label $y$ is one of the categories assigned to doc, chosen uniformly at random if more than one category was assigned to doc.

We compared the CPT to the one-against-all algorithm, a standard approach for reducing multi-class regression to binary regression. The one-against-all approach regresses on the probability of each category $c$ versus all other categories. Given a base training example $(x, y)$, the example used to train the regressor $f_c$ for category $c$ is $(x, I[y = c])$, where $I[\cdot]$ is the indicator function. Predictions for a new test example $(x, y)$ are done according to $f_y(x)$. The learning algorithm used for training the binary regressors in both approaches was incremental gradient descent with squared loss. For each algorithm, we ran several versions with different learning rates, chosen from a coarse grid, and picked the setting that yielded the smallest training error. For the CPT algorithm, we performed a similar search over $\alpha$.

The one-against-all approach used one pass over the training data, while the CPT used two passes. Note that even with an additional pass, the CPT is much faster than one-against-all for training, due to the fact that CPT requires training only about log(number of categories) = log(103) regressors (nodes in the tree) per example, whereas one-against-all trains one regressor per category. On our machine, the CPT took 108 seconds to train, while one-against-all took 2300 seconds. We use Progressive Validation [1] to compute an average squared loss over the test set with results appearing in the following table, where the confidence intervals are computed by Hoeffding's inequality [3] with $\delta = 0.05$.

| One-against-all | $0.55 \pm .012$ |
|---|---|
| CPT with a random tree | $0.56 \pm .012$ |
| CPT with a balanced tree | $0.56 \pm .012$ |
| CPT with an online tree ($\alpha = 0.6$) | $0.56 \pm .012$ |

The values are indeed mostly identical, but CPT achieved this performance with an order of magnitude less computation.

Note that in this problem, there is not much advantage in using our algorithm over using a random tree. Since there aren't many labels and there are many examples, the structure of the tree is not very important. This is confirmed by running the algorithm with various different random trees and observing little variability in squared loss.

### 4.3.2 Web Advertising

We used a proprietary dataset consisting of about $50M$ pairs of webpages and associated advertisements that were shown on the webpage. There are about $5.8M$ unique webpages and $860K$ unique ads in the dataset. The most frequent ad appeared in approximately 1.2% of the cases. The events were split into a training set of size $40M$, and a test set of size $10M$ in time order. Note that webpages and ads both appear multiple times in the training and test sets. For each event, where an event consisted of a single ad being shown on a single webpage, we create a sample $(x, y)$, where $x$ is a "bag of words" vector representation of the webpage, and $y$ is a unique ID associated with the advertisement displayed. The learning problem is predicting $P(y \mid x)$, or the probability that the logging policy displays advertisement $y$ given webpage $x$. Since $n$ is large, one-against-all would be extremely slow. The running time for our algorithm on this dataset was about 60 minutes. Multiplying by $860k/\log_2(860k)$ suggests a running time for one-aginst-all of about 5 years.

Besides the three versions of CPT described above, we tested one other method we call the "table-based" method. In the table-based method, we simply predict $P(y \mid x)$ by the empirical frequency with which ad $y$ was displayed on webpage $x$ in the training set. The progressive validation [1] results of the four algorithms over the test set appear in the following table with confidence intervals again computed using Hoeffding's bound for $\delta = 0.05$.

| Method | Squared Loss | Equivalent |
|---|---|---|
| Table | $0.812 \pm .00055$ | 10.11 |
| Random tree | $0.7742 \pm .00055$ | 8.32 |
| Balanced tree | $0.7725 \pm .00055$ | 8.25 |
| Online tree ($\alpha = 0.9$) | $0.7632 \pm .00055$ | 7.91 |
| Best possible | 0.665 | 5.42 |

Here, the "Equivalent" column is the number of labels for which a uniform random process produces the same loss. The "Best possible" line is an unachievable bound on performance found by examining the empirical frequency of ad-webpage pairs in the test set.

The magnitude of squared loss improvement is modest, but substantial enough to be useful. Since many of the webpages are seen many times, the conditional distribution over ads can be approximated well by empirical frequencies. Thus, the table-based method forms a strong baseline. A small but significant fraction of the webpages were seen only a few times, and for these webpages, it was necesssary to generalize (predict which ads would appear based on which ads appeared on pages similar to the current one). On these examples, the tree performed substantially better.